\documentclass{tlp}

\usepackage[utf8]{inputenc}
\usepackage{amsmath}
\usepackage{amssymb}
\usepackage{microtype}
\usepackage{url}
\usepackage{graphicx}

\usepackage{hyphenat}
\usepackage[inline]{enumitem}
\usepackage{url}
\usepackage[hidelinks]{hyperref}
\usepackage{listings}
\usepackage{csquotes}
\usepackage{algorithmicx}
\usepackage[linesnumbered,ruled,vlined]{algorithm2e}

\usepackage{pgf}

\usepackage{listings}

\providecommand{\Underscore}{\textunderscore}

\lstdefinelanguage{clingo}{basicstyle=\ttfamily,keywordstyle=[1]\bfseries,keywordstyle=[2]\bfseries,keywordstyle=[3]\bfseries,showstringspaces=false,literate={_}{\Underscore}1 {\%\%}{}0,escapeinside={\#(}{\#)},alsoletter={\#,\&},keywords=[1]{not,from,import,def,if,else,elif,return,while,break,and,or,for,in,del,and,class,with,as,is,yield,async},keywords=[2]{\#const,\#show,\#minimize,\#base,\#theory,\#count,\#external,\#program,\#script,\#end,\#heuristic,\#edge,\#project,\#show,\#sum},keywords=[3]{&,&dom,&sum,&diff,&show},morecomment=[l]{\#\ },morecomment=[l]{\%\ },morestring=[b]",stringstyle={\itshape},commentstyle={\color{darkgray}}}

\lstdefinelanguage{python}{basicstyle=\ttfamily,keywordstyle=[1]\bfseries,showstringspaces=false,literate={_}{\Underscore}{1},escapeinside={\#(}{\#)},alsoletter={\#,\&},keywords=[1]{not,from,import,def,if,else,elif,return,while,break,and,or,for,in,del,and,class,with,as,is,yield,async},morecomment=[l]{\#\ },morestring=[b]",stringstyle={\itshape},commentstyle={\color{darkgray}}}

\sbox0{\ttfamily A}
\edef\mybasewidth{\the\wd0 }
\sbox0{\scriptsize\ttfamily A}
\edef\mybasewidths{\the\wd0 }

\lstdefinelanguage{clingos}{language=clingo,
	columns=flexible,
	basewidth=\mybasewidths,
	basicstyle=\scriptsize\ttfamily
}
\lstdefinelanguage{clingoht}{language=clingo,
	columns=flexible,
	basewidth=\mybasewidth,
	escapeinside=||,
	mathescape,
	float=ht,
}

\newtheorem{definition}{Definition}

\providecommand{\sysfont}{\textit}

\newcommand{\clingraph}{\sysfont{clingraph}}
\newcommand{\clinguin}{\sysfont{clinguin}}

\newcommand{\clingo}{\sysfont{clingo}}

\newcommand{\pc}{\ensuremath{\mathcal{P}}}
\newcommand{\pcc}{\ensuremath{\mathcal{C}}}

\newcommand{\Objectneeded}{\textit{ObjectNeeded}}
\newcommand{\Gbupper}{\textit{GlobalUpperBoundGap}}
\newcommand{\Lbupper}{\textit{GlobalLowerBoundGap}}
\newcommand{\Associationpossible}{\textit{AssociationPossible}}

\newcommand{\cpclasses}{\ensuremath{\mathit{C}}}
\newcommand{\cpassocnames}{\ensuremath{\mathit{AN}}}
\newcommand{\cpassocname}{\ensuremath{\mathit{an}}}
\newcommand{\cpassocs}{\ensuremath{\mathit{A}}}
\newcommand{\cpattrnames}{\ensuremath{\mathit{AtN}}}
\newcommand{\cpattrs}{\ensuremath{\mathit{At}}}
\newcommand{\cpattr}{\ensuremath{\mathit{at}}}
\newcommand{\cpdomain}{\ensuremath{\mathit{D}}}
\newcommand{\cpobjects}{\ensuremath{\mathit{O}}}

\newcommand{\confsatconstr}{\ensuremath{\mathit{sat}}}
\newcommand{\confusedobjs}{\ensuremath{\mathit{O}}}
\newcommand{\confinstanceof}{\ensuremath{\mathit{instanceof}}}
\newcommand{\confassocs}{\ensuremath{\mathit{assocs}}}
\newcommand{\confvalues}{\ensuremath{\mathit{values}}}

\begin{document}

\lefttitle{L.\ Bal\'a\v{z}ov\'a, R.\ Comploi-Taupe, S.\ Hahn, N.\ R\"uhling, G.\ Schenner}

\jnlPage{1}{8}
\jnlDoiYr{2021}
\doival{10.1017/xxxxx}

\title[Smart Expansion Techniques for ASP-based Interactive Configuration]{Smart Expansion Techniques for ASP-based Interactive Configuration
\thanks{This research was partially supported by the Austrian Research Promotion Agency (FFG) under the \enquote{AI for Green} programme, and by the German Federal Ministry for Economic Affairs and Energy (BMWE) through the ZIM project grant KK5291307GR4.}
}

\begin{authgrp}
	\author{\sn{Lucia} \gn{Bal\'a\v{z}ov\'a}}
	\affiliation{SIEMENS AG \"Osterreich, Austria}
	\author{\sn{Richard} \gn{Comploi-Taupe}}
	\affiliation{SIEMENS AG \"Osterreich, Austria}
	\author{\sn{Susana} \gn{Hahn}}
	\affiliation{University of Potsdam, Germany}\affiliation{Potassco Solutions, Germany}
	\author{\sn{Nicolas} \gn{R\"uhling}}
	\affiliation{University of Potsdam, Germany}\affiliation{UP Transfer, Germany}
	\author{\sn{Gottfried} \gn{Schenner}}
	\affiliation{SIEMENS AG \"Osterreich, Austria}
\end{authgrp}

\maketitle

\begin{abstract}
	Product configuration is a successful application of Answer Set Programming (ASP). However, challenges are still open for interactive systems to effectively guide users through the configuration process.
	The aim of our work is to provide an ASP-based solver for interactive configuration that can deal with large-scale industrial configuration problems and that supports intuitive user interfaces via an API.
	In this paper, we focus on improving the performance of automatically completing a partial configuration.
	Our main contribution enhances the classical incremental approach for multi-shot solving by four different smart expansion functions.
The core idea is to determine and add specific objects or associations to the partial configuration by exploiting cautious and brave consequences before checking for the existence of a complete configuration with the current objects in each iteration.
This approach limits the number of costly unsatisfiability checks and reduces the search space, thereby improving solving performance.
	In addition, we present a user interface that uses our API and is implemented in ASP.\\
	\textit{Under consideration for acceptance in TPLP.}
\end{abstract}


\section{Introduction}

Configuration~\citep{kbc14} has been one of the first successful applications~\citep{gekasc11c,fefaateruraz17a,gescer19a}
of Answer Set Programming (ASP; \cite{lifschitz19a,gelkah14a}).
Nonetheless, there are open challenges for ASP-based configurators,
one of them being interactive configuration.

Industrial configuration problems in large infrastructure projects may contain
thousands of components and hundreds of component types and
are typically solved in a step-wise manner
by combining interactive actions with automatic solving of sub-problems \citep{fahakrscscta20a}.
When using a grounding-based formalism like ASP,
a grounding bottleneck \citep{eifafiwo07a} arises
due to the large number of components. Furthermore, the necessary domain size is not known beforehand
and can vary significantly.
Therefore, we require a way to dynamically introduce new components during the configuration process.

In our previous work we developed an API to satisfy basic requirements for interactive configuration \citep{fahakrscscta20a}
based on OOASP~\citep{faryscsh15a} and
using various features of \clingo\footnote{\url{https://potassco.org/clingo/}} \citep{gekakasc17a}. In this paper, we report on our recent advancements in developing an ASP-based domain-independent interactive configuration platform.
The new achievements, compared to previously published reports \citep{cohascsc22a,cofahascsc23a}, consist of dramatically improved solving performance
by using novel so-called \emph{smart expansion} techniques and the creation of a UI prototype based on \textit{clinguin} \citep{behasc24a}.

The general idea of the smart expansion techniques is to infer as much knowledge as possible from the current configuration state
and to use this knowledge to reduce the search space of the ASP solver and to limit costly unsatisfiability checks.
This results from the insight that in our earlier implementation objects were added incrementally to the configuration without any class
restrictions or associations which in turn led to a combinatorial explosion of the search space.
For this purpose, we introduced four new functions: \Objectneeded, \Gbupper, \Lbupper, and \Associationpossible.
While the first three ones use intersections of possible solutions to determine objects that are definitely needed in the configuration,
the latter function provides possible associations between two objects with the intention of quickly estimating the minimal size of a complete configuration.
We demonstrate the working of the functions using our racks example as known from earlier work.

The remainder of the paper is organized as follows:
Section~\ref{sec:background} presents background on ASP, product configuration and the OOASP framework,
as well as our running example and previous results regarding the interactive configurator.
We then proceed to present our novel contributions. Section~\ref{sec:contributions} contains
detailed explanations about the four smart expansion functions and their implementation in our interactive configurator,
and Section~\ref{sec:ui} introduces our novel \textit{clinguin} user interface.
Finally, Section~\ref{sec:experimentalresults} demonstrates the experimental results of our new implementation
comparing its performance to the previous version and Section~\ref{sec:conclusions} concludes this paper.
 \section{Background}\label{sec:background}

\subsection{Answer Set Programming}\label{sec:background:asp}

A {logic program} consists of {rules} of the form
\begin{lstlisting}[mathescape,numbers=none]
   a$_1$;...;a$_m$ :- a$_{m+1}$,...,a$_n$,not a$_{n+1}$,...,not a$_o$.
\end{lstlisting}
where for $1 \leq i \leq o$, each \lstinline[mathescape]{a$_i$} is
an {atom} of form \lstinline[mathescape]{p(t$_1$,...,t$_k$)}
and all \lstinline[mathescape]{t$_j$} are terms
(variables, constant terms, or function terms).
For $0 \leq m \leq n \leq o$,
atoms \lstinline[mathescape]{a$_1$} to \lstinline[mathescape]{a$_m$} are often called head atoms,
while \lstinline[mathescape]{a$_{m+1}$} to \lstinline[mathescape]{a$_n$}
    and \lstinline[mathescape]{not a$_{n+1}$} to \lstinline[mathescape]{not a$_o$}
are also referred to as positive and negative body literals, respectively.
An expression is said to be {ground} if it contains no variables.
As usual, \lstinline[mathescape]{not} denotes (default) {negation}.
A rule is called a {fact} if $m=n=o=1$,
normal if $m=1$, and
an integrity constraint if $m=0$.
In what follows, we deal with normal logic programs only,
for which $m$ is either 0 or 1.
Semantically, a logic program induces a set of {stable models},
being distinguished models of the program determined by the stable models semantics~\citep{gellif90a}.

To ease the use of ASP in practice,
several extensions have been developed.
Multi-shot solving, which is one such extension, allows for solving continuously changing logic programs in an operative way.
In \clingo,
this can be controlled via an API for implementing reactive procedures that loop on grounding and solving while reacting,
for instance, to outside changes or previous solving results.

We want to highlight here the use of so-called \emph{assumptions} and \emph{externals}~\citep{karoscwa21a}.
The former are added to the solving process and can be interpreted as integrity constraints that force
the answer sets to contain certain atoms without providing evidence for them.
The latter are specified by the \lstinline{#external} directive and allow for declaring atoms whose truth value can be set via the API.
This provides the tools to continuously assemble ground rules evolving at different stages of a reasoning process
and to change program behavior by manipulating the truth values of atoms.

 \subsection{Product Configuration and OOASP}\label{sec:background:configuration}

Product configuration as an activity produces the specification of an artifact that is assembled from instances of given component types and that conforms to a given set of constraints between those components.
Component types can have attributes, thus components can be parametrized.
Furthermore, components are related via part-of, is-a, or other relationships \citep{kbc14}.
Many configuration problems are dynamic, meaning the number of necessary components for a solution is unknown in advance \citep{fafrhascsc16a}.

OOASP \citep{faryscsh15a,fafrsctate18a} is an ASP-based framework to encode and reason about object-oriented problems such as configuration problems.
It defines a Domain Description Language (DDL) specific to the domain of object-oriented models that can be represented by a modelling language corresponding to a UML class diagram.
OOASP-DDL defines ASP predicates to encode models (classes, subclass relations, associations, and attributes) and instantiations (instances, is-a relations, instance-level associations, and attribute values).
Furthermore, it provides a uniform way to encode constraints.
Table~\ref{tab:ooasp:models} shows the OOASP-DDL predicates for the encoding of models, and Table~\ref{tab:ooasp:instantiations} shows the OOASP-DDL predicates for the encoding of instantiations.\footnote{We here present a version of OOASP-DDL that has evolved from the original definition \citep{faryscsh15a} and that has also been slightly simplified for this paper.}

\begin{table*}
	\caption{OOASP-DDL predicates for the encoding of models}
	\label{tab:ooasp:models}
	\begin{tabular}{|l|l|}
	\cline{1-2}
	\textbf{Predicate} & \textbf{Description} \\ \cline{1-2}
	\lstinline|ooasp_class(C)| & \lstinline|C| is a class \\ \cline{1-2}
	\lstinline|ooasp_subclass(SubC,SupC)| & \lstinline|SubC| is a subclass of \lstinline|SupC| \\ \cline{1-2}
	\lstinline|ooasp_assoc(A,C1,C1Min,C1Max,| & In association \lstinline|A|, each instance of class \lstinline|C1| \\
\lstinline|             C2,C2Min,C2Max)| & is associated to \lstinline|C2Min|-\lstinline|C2Max| instances of class \lstinline|C2|,\\
& and each \lstinline|C2| instance to \lstinline|C1Min|-\lstinline|C1Max| \lstinline|C1| instances \\ \cline{1-2}
	\lstinline|ooasp_attr(C,A,T)| & \lstinline|A| is an attribute of class \lstinline|C| with type \lstinline|T| \\ \cline{1-2}
	\lstinline|ooasp_attr_enum(C,A,D)| & \lstinline|D| is an element of the domain of attribute \lstinline|A| of class \lstinline|C| \\ \cline{1-2}
	\end{tabular}
\end{table*}

\begin{table*}
	\caption{OOASP-DDL predicates for the encoding of instantiations}
	\label{tab:ooasp:instantiations}
	\begin{tabular}{|l|l|}
	\cline{1-2}
	\textbf{Predicate} & \textbf{Description} \\ \cline{1-2}
		\lstinline|ooasp_isa(C,O)|            & \lstinline|O| is an object of class \lstinline|C| \\\cline{1-2}
\lstinline|ooasp_associated(A,O1,O2)| & Object \lstinline|O1| is associated to object \lstinline|O2| in association \lstinline|A| \\\cline{1-2}
		\lstinline|ooasp_attr_value(A,O,V)| & The attribute \lstinline|A| of object \lstinline|O| has value \lstinline|V|\\ \cline{1-2}
	\end{tabular}
\end{table*}

OOASP constraints are defined using the predicate \lstinline|ooasp_cv| (where \enquote{cv} stands for \enquote{constraint violation}).
Rules with head atoms of this predicate are used instead of ASP constraints to enable configurations to be \emph{checked}, i.e., to derive which constraints are violated in a given configuration.
To enforce a configuration to be consistent, the ASP constraint: \lstinline|:- ooasp_cv(C,O,M,L)| is added, forbidding any constraint violations.
An \lstinline|ooasp_cv| atom contains four terms: a unique constraint identifier \lstinline|C|, the identifier of the faulty object \lstinline|O|, a string containing a message \lstinline|M| describing the issue, and a list \lstinline|L| of additional explanatory terms.

OOASP distinguishes integrity constraints from domain-specific constraints.
The former are defined in the OOASP framework itself and refer to issues such as invalid values and violations of association cardinalities.
Domain-specific constraints can be defined by a user of OOASP as additional rules that derive \lstinline|ooasp_cv| atoms.

An instantiation (configuration) defined by the predicates from Table~\ref{tab:ooasp:instantiations} is complete if every object is an instance of an instantiable class, and it is valid (consistent) if no constraint violations can be derived from it.
We follow the convention that only leaf classes (i.e., classes that have no subclasses) are instantiable, so every object must be an instance of a leaf class in a complete configuration.
Finding a complete and valid configuration is usually an interactive task, iteratively involving user interactions (decisions) and automatic reasoning by a solver, e.g., an ASP solver \citep{fahakrscscta20a}.
The goal of our work is to support interactive configuration in a framework based on OOASP.
 \subsection{Running example}\label{sec:example}

\begin{figure*}[h]
    \centering
    \includegraphics[width=\textwidth]{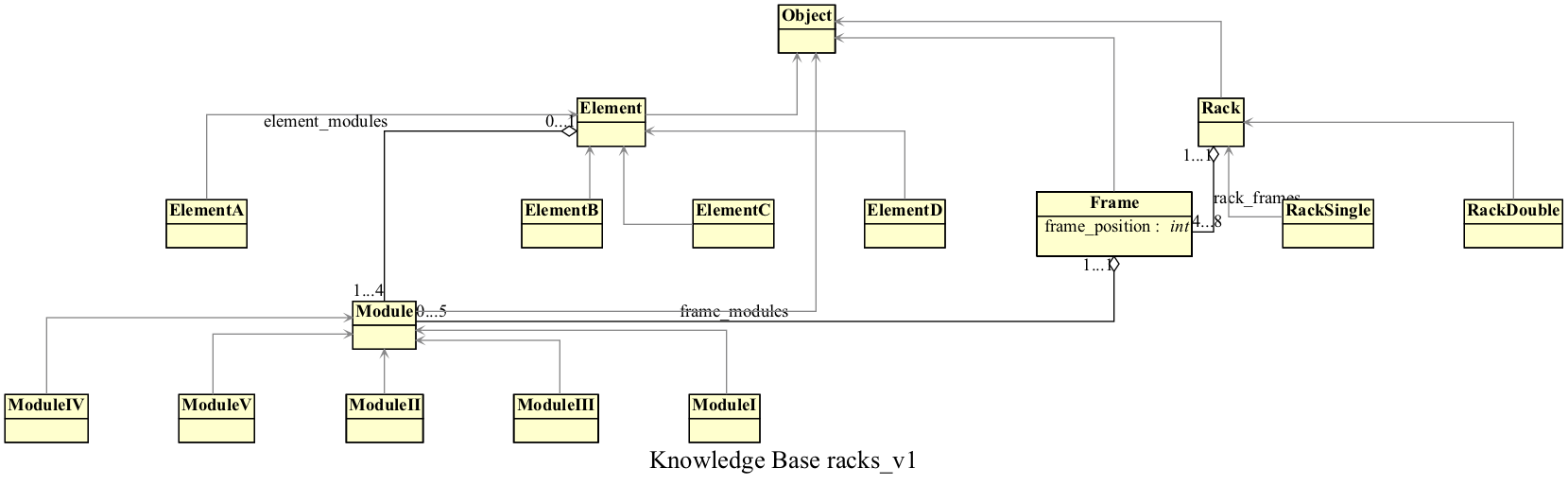}
    \caption{Class diagram for the racks knowledge base generated by \clingraph.}
    \label{fig:uml-diagram}
\end{figure*}

We use an extension of the typical hardware racks configuration problem \citep{faryscsh15a} as running example.
The UML class diagram (Figure~\ref{fig:uml-diagram}) shows all concepts and relations of the racks knowledge base.
Additionally,
it includes the following domain-specific constraints:
\begin{enumerate}
    \item[C1.] An ElementA/B/C/D requires exactly 1/2/3/4 objects of type ModuleI/II/III/IV
    \item[C2.] Instances of ModuleI/II/III/IV must be connected to exactly one Element
    \item[C3.] A RackSingle/RackDouble has exactly 4/8 Frames
    \item[C4.] A Frame containing a ModuleII must also contain exactly one ModuleV
\end{enumerate}

This example captures the essence of a typical configuration knowledge base in an industrial setting.
Of course, real-life industrial knowledge bases are much larger ($>$100 classes, associations, attributes).
Another property of these knowledge bases is that the number of objects required for a solution is not known beforehand.
 \subsection{Interactive Configurator}\label{sec:approach}

In our previous work \citep{cofahascsc23a}\footnote{Source code for this version can be found in \url{https://github.com/siemens/OOASP/tree/v1.0.0}}, we have
identified and implemented eight distinct interactive tasks for product configuration.
We have also provided an API as well as a user interface implemented with \textit{ipywidgets}
and using \textit{clingraph},
to visualize the model and the configuration graph.
The interactive configurator presented in this paper
enables users to build a complete configuration incrementally based on the OOASP framework,
by combining user actions with automatic solving.
The implementation takes advantage of \clingo's multi-shot capabilities \citep{gekakasc17a},
ensuring that rules are grounded dynamically as the configuration evolves.
This approach avoids the need for re-grounding and utilizes learned constraints.

The system allows users to carry out edition tasks (T1 to T4)
which amount to setting the type of existing objects,
adding or removing associations,
setting values for attributes, and
adding new objects.
Additionally, three key reasoning tasks are integrated into the system:
task T5 generates the configuration using choice rules,
task T6 checks if the configuration violates any constraints,
and task T7 provides the user with available edit options through brave reasoning.
Tasks T6 and T7 employ two externals, namely,
\lstinline|check_permanent_cv| and \lstinline|check_potential_cv|,
which help manage constraint violations.
The potential violations are those that can be fixed later in the process,
while permanent violations cannot be resolved.
Finally, task T8 encapsulates the overall process,
where a partial configuration ($\pc$) is extended incrementally into a complete and consistent one ($\pcc$),
by adding new objects until a complete and consistent configuration is found.
This task was identified as the most expensive one in terms of performance,
as it requires the solver to check for the existence of $\pcc$ with the current objects
leading to multiple costly unsatisfiability checks before finding a solution.

\section{Improved interactive configuration}\label{sec:contributions}

In this section we outline the new contributions
that build upon the work presented by \cite{cofahascsc23a} and
whose main goal is to improve the performance and usability of the interactive configurator.

Let us first provide a novel formalisation of task T8, which will be the focus of our improvements.
\begin{definition}
	A \textbf{configuration problem} is defined by a tuple $\langle \cpclasses, \cpassocnames, \cpassocs, \cpattrnames, \cpattrs, \cpdomain, \cpobjects, \confsatconstr \rangle$,
	where $\cpclasses$ is the set of instantiable classes,
	$\cpassocnames$ is a set of unique association names,
	$\cpassocs: \cpassocnames \rightarrow \cpclasses \times \cpclasses$ is a total function that defines the involved classes for each association,
	$\cpattrnames$ is a set of attribute names,
	$\cpattrs \subseteq \cpclasses \times \cpattrnames$ is the set of attributes,
	$\cpdomain$ is the universal domain of attribute values,
	$\cpobjects$ is the (possibly infinite) set of potentially usable objects,
	and $\confsatconstr$ is a function that maps a partial configuration $\pc$ as defined below to the value $\top$ iff $\pc$ satisfies all the constraints imposed by the configuration problem.\footnote{The detailed way how to define constraints is abstracted away here. Constraints include integrity constraints as well as domain-specific constraints as described in Section~\ref{sec:background:configuration}.}
\end{definition}
\begin{definition}
	A \textbf{(partial) configuration} $\pc$ of a configuration problem as defined above is defined by a tuple $\langle \confusedobjs_\pc, \confinstanceof_\pc, \confassocs_\pc, \confvalues_\pc \rangle$,
	where $\confusedobjs_\pc \subseteq \cpobjects$ is the set of instantiated objects,
	$\confinstanceof_\pc: \confusedobjs \rightharpoonup \cpclasses$ is a partial function mapping instantiated objects to their classes,
	$\confassocs_\pc: \cpassocnames \rightarrow 2^{\confusedobjs_\pc \times \confusedobjs_\pc}$ is a function defining the instantiated associations,
	and $\confvalues_\pc: \cpattrs \times \confusedobjs \rightharpoonup \cpdomain$ is a partial function mapping attributes to their values.
\end{definition}
A configuration is \emph{complete} if every object is an instance of an instantiable class (i.e., the function $\confinstanceof$ is total) and every attribute has a value (i.e., for all $\cpattr = (c,a) \in \cpattrs$ and all $o \in \confusedobjs_\pc$ where $\confinstanceof_\pc(o) = c$, we get that $\confvalues_\pc(\cpattr,o)$ has a value), and it is \emph{consistent} if the constraints are satisfied (i.e., $\confsatconstr(\pc) = \top$).

The latter condition also enforces the lower bounds of all associations to be satisfied, and constraints may restrict the domains of attributes.

Task T8, extending a partial configuration into a complete and consistent one, can then be formally defined as follows:
An incremental extension of a partial configuration to a complete and consistent one is
a sequence of (partial) configurations $\langle \pc_1, \dots, \pc_n \rangle$,
where for each $1 \leq i < n$, we get that $\pc_{i+1}$ is a proper extension of $\pc_i$, i.e., it holds that $\pc_i \neq \pc_{i+1}$,
$\confusedobjs_{\pc_i} \subseteq \confusedobjs_{\pc_{i+1}}$,
$\confinstanceof_{\pc_i} \subseteq \confinstanceof_{\pc_{i+1}}$,
$\confvalues_{\pc_i} \subseteq \confvalues_{\pc_{i+1}}$,
and for all $\cpassocname \in \cpassocnames: \confassocs_{\pc_i}(\cpassocname) \subseteq \confassocs_{\pc_{i+1}}(\cpassocname)$.
Furthermore, the configuration $\pc_n$ is complete and consistent as defined above.

\paragraph*{}

Our main contribution in this paper is the introduction of so-called
\emph{smart expansion functions}.
For this purpose, we present four different functions
which focus on improving the performance of task T8
by reducing the search space of the configuration problem.
In the most basic implementation of task T8,
objects are added incrementally to the configuration without any class restrictions or associations,
leading to a combinatorial explosion of the search space.
To mitigate this,
we inspected new techniques that allow us to use existing information
to infer additional knowledge about the required objects
while also reducing the number of unsatisfiability checks.

We enhanced the usability of the system by extending the OOASP-DDL
with the concept of association specializations \citep{tighterbounds}.
This concept mirrors the idea of subclassing in OOASP but applies it to associations.
With this extension,
users can define associations specific to subclasses that override those of their parent classes.
Therefore, domain-specific constraints C1-C3 from Section~\ref{sec:example} can now be integrated into the knowledge base directly.
For instance, by adding the facts
\lstinline{ooasp_assoc_specialization(element_modules1, element_modules)} and
\lstinline{ooasp_assoc(element_modules1, elementA, 1, 1, moduleI, 1, 1)},
we introduce a specialized version of the \lstinline{element_modules} association, namely
\lstinline{element_modules1}, which enforces that each \lstinline{elementA} must be associated
with exactly one \lstinline{moduleI}, and vice versa.
However, constraint C4 cannot be expressed in this way and remains directly implemented in the ASP encoding.
Overall, this extension increases the expressiveness of the language,
reduces the need for domain-specific constraints,
and enables the encoding to leverage this structural information during solving,
as well as for the inferences in the smart functions.

Additionally, we also improved the performance of the system
by removing the use of externals for user input and replacing them with assumptions
which was possible thanks to the reframing of task T6 (checking if $\pc$ is complete or if it violates any constraints).
Originally, this task needed to generate facts
about which constraints were violated without solving the configuration.
For this, it was necessary to disable all choice rules to avoid fixing the violations.
However,
since assumptions can be considered as constraints,
using them to represent user input
is only possible if the program has rules to obtain the assumed atoms,
which requires the choice rules.
We were able to refocus this in our new version
by restricting feedback about constraint violations to use the smart functions described below.
This change omits certain feedback details, such as missing attributes,
but still provides information about missing objects and associations, which is arguably more relevant,
at the benefit of removing the complexity introduced by externals.

We dedicate the remainder of this section to the illustration of our implementation of task T8
followed by a detailed explanation of the four smart expansion functions.

\subsection{Smart functions}

As mentioned in Section~\ref{sec:approach}, our basic approach to solving the configuration problem incrementally
consisted in adding objects to the configuration one by one
and checking for unsatisfiability after each addition.
Our novel contribution enhances this multi-shot approach by adding four different smart expansion functions.
The idea is that before checking for the existence of $\pcc$ with the current (grounded) objects,
we apply the smart expansion functions, which determine and add necessary specific objects or associations to $\pc$.
This postpones costly unsatisfiability checks to the end of the process
and thereby limits their number to a minimum.
As a consequence, satisfiable instances might be processed by the smart expansion functions
but since they already meet the constraints, no new objects will be added.
The overhead of this additional computation is minimal compared to the cost of checking unsatisfiability after adding each object.

The entire process is outlined in Algorithm~\ref{alg:incremental_configuration},
which takes a partial configuration $\pc$ and iterates until it is extended into a complete configuration.
It first sets the external \texttt{check\_potential\_cv} to false,
allowing the smart expansion functions to analyze the current configuration.
Then, it iterates over the smart functions,
which are responsible for adding new objects or associations to the partial configuration.
Selection and order of smart functions can be customized.
If any of the functions add new objects or associations,
the algorithm starts again with the first smart function.
Once $\pc$ can't be extended further by the smart functions,
the external \texttt{check\_potential\_cv} is set to true,
and the solver is called to check for unsatisfiability.
If the configuration is still unsatisfiable,
it means that the current set of objects is insufficient to satisfy the constraints
and that the smart functions were not able to detect this.
In this case, the algorithm adds a new object without a fixed type
to the configuration.
Throughout this process, the program is grounded incrementally
whenever a new object is added, either by the smart functions or by the algorithm itself.

\begin{algorithm}
	\caption{Smart Incremental Solving}
	\label{alg:incremental_configuration}
	\KwIn{Partial configuration $\pc$}
	done $\gets$ \textbf{False}\;
	\While{not \textnormal{done}}{
		Set external \texttt{check\_potential\_cv} to \textbf{False}\; \label{alg:incremental_configuration:jumpto}
\ForEach{f $\in$ \texttt{smart\_functions}}{
			config\_was\_extended $\gets$ $f(\pc)$\;
			\If{\textnormal{config\_was\_extended}}{
				\textbf{goto} line \ref{alg:incremental_configuration:jumpto}\;
			}
		}
		Set external \texttt{check\_potential\_cv} to \textbf{True}\;
		done $\gets$ Solve()\;
		\If{not \textnormal{done}}{
			Add abstract object\;
		}
	}
\end{algorithm}

The smart expansion functions exploit knowledge about the configuration model
using auxiliary predicates to gather relevant information about the current configuration state.
These atoms are then used to determine necessary objects and associations
by calculating cautious and brave consequences (intersection and union of stable models, respectively).
To ensure a satisfiable answer that provides this information,
we ignore potential constraints by setting the external \lstinline|check_potential_cv| to false,
while the permanent constraints remain active,
since we want to discard anything that cannot be fixed by further interaction with the system.

We now proceed to present each smart expansion function in detail.
The encodings presented here are simplified versions that omit the additional argument used for incrementally grounded objects.
For the complete encodings, we refer the reader to the source code.\footnote{\url{https://github.com/siemens/OOASP/tree/v2.0.0}}

\subsubsection{\Objectneeded}

The first smart expansion function detects when additional objects of a specific class are required, creates them and associates them with an existing object.

To illustrate the workings of the function, we will consider the following example in the remainder of this section:
The input partial configuration consists of one Rack $r$ and one Frame $f$ without associations.
The model (Figure~\ref{fig:uml-diagram}) specifies that a Rack must be associated with at least four Frames.
This means that even if $r$ gets associated with $f$,
at least three more Frames must be added to satisfy the requirement.

This information is encoded using atoms of the form \lstinline{assoc_needs_object(ID,A,X,C,DIR)},
as shown in Listing~\ref{prg:assocneedsobject}.
This predicate indicates that the object \lstinline{ID}
requires \emph{at least} \lstinline{X} objects of class \lstinline{C},
associated through \lstinline{A} in the direction \lstinline{DIR}.
The rule relies on the predicate \lstinline|lb_violation|, which detects violations of the lower bound for association \lstinline|A|,
where the required number of objects is \lstinline|CMIN| and the current number is \lstinline|N|.
This violation is then used in the head of the rule to determine how many additional objects are needed,
by computing the range from 1 to the missing amount, specifically \lstinline{CMIN-N}.
\begin{lstlisting}[caption={Encoding to obtain the \lstinline{assoc\_needs\_object/6} predicate.}, label={prg:assocneedsobject},language=clingo]
assoc_needs_object(ID,A,1..CMIN-N,C,D) :-
	lb_violation(ID,_,(A,CMIN,N,C,D)).
\end{lstlisting}

The smart function analyzes these atoms within the cautious consequences to determine the necessary objects and associations.
Since these atoms appear in the cautious consequences, they hold true in all models,
indicating that the constraints cannot be satisfied with the current set of objects.
In our example, this analysis leads to the conclusion that at least three additional Frames must be associated with $r$ (identified by ID $1$),
as captured by the following target atoms:
\begin{lstlisting}
assoc_needs_object(1,rack_frames,1,frame,1),
assoc_needs_object(1,rack_frames,2,frame,1),
assoc_needs_object(1,rack_frames,3,frame,1).
\end{lstlisting}

We employ multiple atoms under the concept of ``at least"
and use model intersections to eliminate target atoms
from models where possible associations with existing objects were not considered.
In our example, since potential constraint violations are not enforced,
there exists a model in which $r$ and $f$ are not associated which
obtains target atoms for $X \in \{1,2,3,4\}$.
However, there is another model where $r$ and $f$ are associated,
so the atom for $X=4$ is absent.
As a result, it is discarded during the intersection process.
This technique significantly improves performance compared to our previous attempts,
where a single atom and an optimization statement were used to minimize constraint violations.

As a result, the smart function then analyzes the target atoms, determines the maximum value of $X$, and grounds three additional Frames, directly associating them with object $r$ through the \lstinline|rack_frames| association.

\subsubsection{\Gbupper}

This smart function calculates required objects using upper bounds
by summing up the number of existing instances of a class
and verifying whether the upper bounds can be satisfied globally.
This is implemented in the encoding in Listing~\ref{prg:globalubgap}, where the presence of the atom
\lstinline|global_ub_gap(C,N)| in the cautious consequences signifies the need to add at least \lstinline{N} objects of type \lstinline{C}.

This function is limited to \enquote{target associations},
in which one side involves a single object.
These associations are collected using the predicate \lstinline|is_target_assoc/2|,
which is incorporated into the rule.
The predicate \lstinline|ooasp_assoc_limit/6| is then used to retrieve the upper bound for the association from the model.
Next, the \lstinline|#count| aggregate is applied
to count the number of objects for each class within the association.
The final line ensures that the upper bound exceeds the number of objects already in the configuration
and performs the necessary calculation to determine how many additional objects are needed.

For example, a partial configuration with one Rack and nine Frames includes the atom \lstinline|global_ub_gap(rack,1)|.
This is calculated from the fact that the upper bound for Racks in the \lstinline|rack_frames| association is 8, while the number of Frames is 9.
As a result, the current Racks are insufficient to associate with all the Frames.
In this particular case, the smart function \Objectneeded\ has no effect,
as there exists a configuration where each Frame is associated with the Rack,
meaning the intersection of violations is empty.

It is important to note that, since this is a global calculation,
no instantiated objects are considered, and therefore, no specific associations can be added.

\begin{lstlisting}[caption={Encoding to obtain the \lstinline{global_ub_gap/2} predicate.}, label={prg:globalubgap},language=clingo]
global_ub_gap(C2,1..((NUM1 + MAX - 1) / MAX)-NUM2):-
    is_target_assoc(A,DIR),
    ooasp_assoc_limit(A,max,DIR2,C2,MAX,C1),
    DIR2!=DIR,
    #count { ID:ooasp_isa(C1,ID) } = NUM1,
    #count { ID:ooasp_isa(C2,ID) } = NUM2,
    NUM2 * MAX < NUM1.
\end{lstlisting}

\subsubsection{\Lbupper}

This function operates similarly to the \Gbupper\ function,
but for lower bounds.
In this case, atoms \lstinline|global_lb_gap(C,N)| in the cautious consequences
indicate the need to add at least \lstinline{N} objects of type \lstinline{C}.

For example, consider a scenario with 2 Racks and 4 Frames.
Although locally, the Racks and Frames appear to have enough objects
to satisfy the lower bound of the \lstinline|rack_frames| association,
the corresponding rule computes that the global lower bound for the two Racks is 8 Frames.
Therefore, the atom \lstinline|global_lb_gap(frame, 4)| indicates the need to ground at least four additional Frames to fulfill this gap.

For brevity, we omit the full encoding which is quite similar to Listing~\ref{prg:globalubgap}.

\subsubsection{\Associationpossible}

The final smart function reduces the search space by identifying viable
associations to complete the configuration through brave consequences.
The inclusion of the atom \lstinline|assoc_possible(A,ID1,ID2)| in the brave consequences signifies
that the association \lstinline{A} between objects \lstinline{ID1} and \lstinline{ID2} can (possibly) be added.
These atoms are calculated in Listing~\ref{prg:assocneeded}.

The rule uses the predicate from the smart function \Objectneeded\
to verify that both objects \lstinline{ID1} and \lstinline{ID2} require an association.
Additionally, the predicate \lstinline{potential_assoc/5} ensures that it is indeed possible to associate the objects.
Finally, we confirm that the classes of the objects are already set, preventing the imposition of a class through the association.

While this approach may affect the completeness of the solution by limiting available options,
it significantly speeds up the process of finding or estimating the minimal domain size.
We can see this improvement in an example with 17 Frames,
where adding these associations reduces the search space.

\begin{lstlisting}[caption={Encoding to obtain the \lstinline{assoc\_possible/3} predicate.}, label={prg:assocneeded},language=clingo]
assoc_possible(A,ID1,ID2):-
	assoc_needs_object(ID1,A,1,_,_),
	assoc_needs_object(ID2,A,1,_,_),
	potential_assoc(A,ID1,ID2,C1,C2),
	user(ooasp_isa(C1,ID1)),
	user(ooasp_isa(C2,ID2)).
\end{lstlisting}

\section{User interface with clinguin}\label{sec:ui}

We have developed a new prototypical user interface (UI) for
smart interactive configuration,
replacing the previous \textit{ipywidgets}-based implementation
by \textit{clinguin} \citep{behasc24a}.
\clinguin\ is a system for generating UIs in ASP.
To achieve this, \clinguin\ employs a set of dedicated predicates to define
layout, style, and functionality of the interface, while handling user-triggered events.
This streamlined design simplifies the specification of continuous user interactions
within an ASP system, specifically \clingo.

\begin{figure}[ht]
    \centering
    \frame{\includegraphics[scale=0.3]{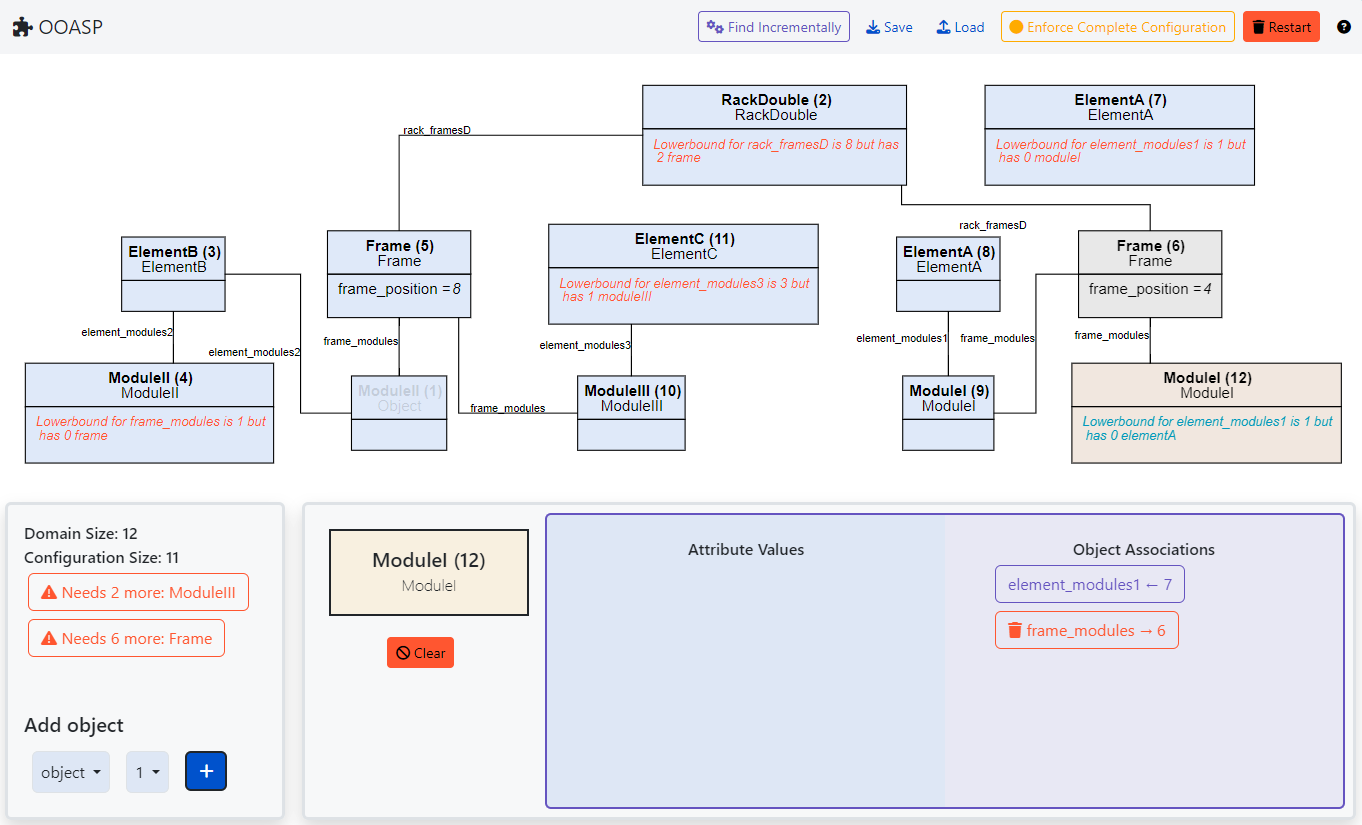}}
    \caption{Snapshot of prototype UI in \clinguin.}
    \label{fig:ui}
\end{figure}

The built-in capacities of \clinguin\ to add user input as assumptions,
and to provide options and inferences via brave and cautious reasoning,
allowed us to directly integrate tasks T1-3, T5, and T7
without the need for additional code.
Our prototype enhances usability with features
such as saving/loading configurations and a \textit{clear} button,
while utilizing \textit{clinguin}'s integration with \clingraph\ \citep{hasascst24a}
to visualize and interact with the configuration graph directly.
A snapshot of the UI is shown in Figure~\ref{fig:ui}.

% --------------------------------------------------------------------------------
% --------------------------------------------------------------------------------
%

 \section{Experimental Results}\label{sec:experimentalresults}
\begin{figure*}[h!]
    \centering
    \resizebox{\textwidth}{!}{\input{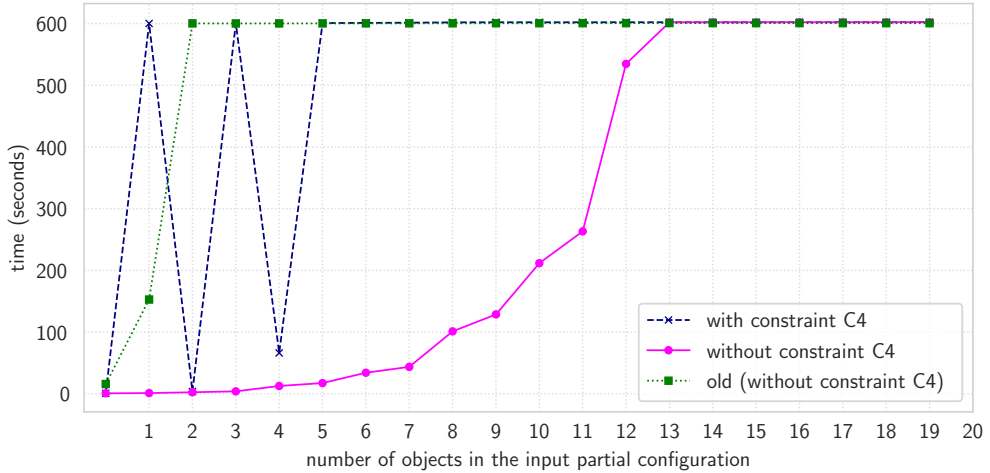}}
    \caption{Runtime (time-out: 600s).}
    \label{fig:bm_runtime}
\end{figure*}

\begin{figure*}[h!]
    \centering
    \resizebox{\textwidth}{!}{\input{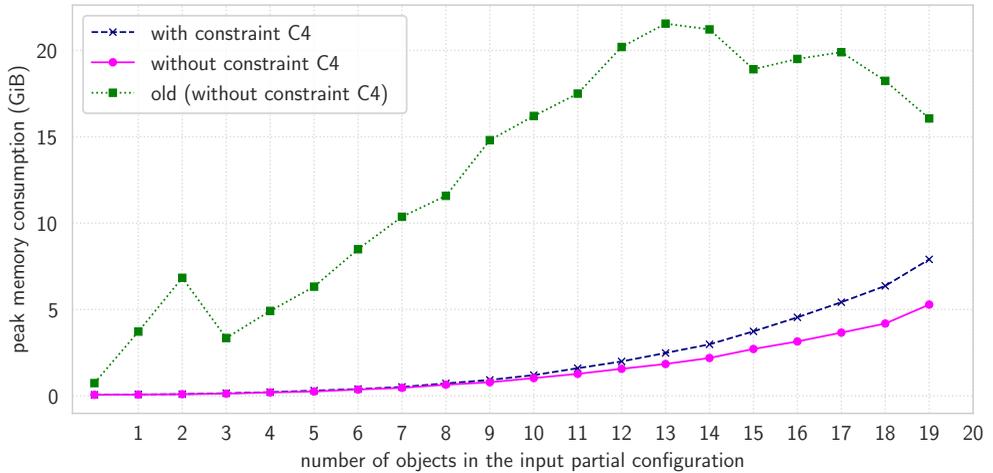}}
    \caption{Peak memory consumption.}
    \label{fig:bm_memory}
\end{figure*}

The performance benchmarks were carried out for task T8 (completing a partial configuration) for 20 problem instances of increasing size and complexity.\footnote{The benchmarking script is available as \texttt{benchmarks/ooasp\_bench.sh} in \url{https://github.com/siemens/OOASP/tree/v2.0.0}}
Each of the instances is defined by a partial configuration containing the same number $n \in 1..20$ of objects of each Element subtype.
The complete configuration obtained for the largest problem instance contains 325 objects.
The performance testing was done on different permutations of the smart functions, with the resource consumption tracked by the \textit{runsolver} tool \citep{roussel_runsolver}.
The testing was performed on a virtual machine with 32GB RAM limit.
Measures were taken to minimize external influence on the performance of the solver, e.g., by disabling swapping.

Figures~\ref{fig:bm_runtime}~and~\ref{fig:bm_memory} visualize our main experimental results in terms of solving performance.
We compare the old version of our implementation \citep{cofahascsc23a} with two different setups of the newest version.
The curves in the figure represent the best-performing smart function permutations, with and without constraint C4 described in Section~\ref{sec:example}, in comparison to the old version.
The smart expansion functions provide a significant improvement in terms of performance compared to the previous version in our benchmarks.
The decrease in runtime and memory consumption allows the system to complete much more complex partial configurations than the previous version.

Constraints which cannot be expressed in the OOASP language (such as constraint C4), however, cannot always benefit from the smart functions.
This is because the smart functions are not able to leverage the knowledge encoded by such constraints to derive missing objects or associations.
The jumps in the runtimes in Figure~\ref{fig:bm_runtime} can be explained by this fact and by observing that there are certain input (partial) configurations
where the system is forced to derive generic objects (without any specific type) as the smart functions are not able to derive the specific types of needed objects.

Given the nature of the smart functions, there can be a huge disparity in the solver performance depending on the chosen order of application.
Thus, the usage of smart functions does not immediately guarantee better performance.
This variability, however, allows the combinations and the order of the smart functions to be tailored to a particular problem, e.g., in portfolio solving \citep[cf.][]{gekakascsczi11a}.
We experimented with various permutations of at least three of the smart functions.
The results show that the order of the smart functions has a significant impact on the performance such that some permutations can solve an instance in a matter of seconds, while others time out.
On average, permutations of all four smart functions starting with \Associationpossible\ provided the best results.
In the results visualized in Figures~\ref{fig:bm_runtime}~and~\ref{fig:bm_memory}, the permutation $\langle$\Associationpossible, \Objectneeded, \Gbupper, \Lbupper $\rangle$ was used as it has shown to have the best performance on the defined benchmarks.

 \section{Conclusions}\label{sec:conclusions}
When dealing with industrial configuration problems, the favoured approach is often an interactive one,
where the user is solving the problem in a step-wise manner while being guided through the configuration process.
As these kinds of problems typically contain thousands of components, solving performance and the design of a clear, intuitive user experience are key.
Our work addresses both of these issues by improving the performance of our existing
interactive configurator~\citep{cohascsc22a,cofahascsc23a},
and by developing a new prototypical UI based on \clinguin.
While the former is achieved by introducing four new \emph{smart expansion functions} to the ASP-based configuration API,
the latter simplifies the specification of continuous user interactions within an ASP system,
thereby facilitating future development of a user-friendly UI.
The four smart expansion functions use cautious and brave reasoning to derive knowledge about the current configuration
and to reduce the search space of the ASP solver.
We demonstrated the working of the functions using our racks example as known from earlier work and compared the performance of the new functions with the old ones.
Our experimental results show that the new functions are able to significantly reduce solving time and memory usage of the ASP solver \clingo.

Future work should explore ways to generalize the smart functions such that they can be applied to configuration problems containing
constraints not specified directly in the OOASP framework.
We also plan to extend the UI prototype to support more complex configuration problems and improve the user experience.

\section*{Competing interests}
The authors declare none.
\bibliographystyle{plainnat}

\end{document}